\documentclass[11pt]{article}
\usepackage{geometry}

\usepackage[english]{babel}
\usepackage{amssymb,amsmath}
\usepackage{array}
\usepackage{amsthm}
\usepackage[symbol]{footmisc}
\usepackage{graphicx}
\usepackage[shortlabels]{enumitem}
\usepackage[dvipsnames]{xcolor} 
\usepackage{algorithm}
\usepackage{algpseudocodex}
\usepackage{authblk}
\usepackage{mathtools}
\usepackage{bbm}
\usepackage{booktabs}
\usepackage{url}

\newcommand{\ee}{\varepsilon}

\newtheorem*{theorem*}{Theorem}

\theoremstyle{definition}

\usepackage{titlesec}
\titlespacing*{\paragraph}{0pt}{0.4\baselineskip}{0.5em}

\title{Persistence-Augmented Neural Networks}

\author[1]{\large Elena Xinyi Wang}
\author[2]{\large Arnur Nigmetov}
\author[2]{\large Dmitriy Morozov}
\affil[1]{\footnotesize Department of Informatics, University of Fribourg}
\affil[2]{\footnotesize Lawrence Berkeley National Laboratory}

\begin{document}
\date{}
\maketitle

\begin{abstract}
Topological Data Analysis (TDA) provides tools to describe the shape of data, but integrating topological features into deep learning pipelines remains challenging, especially when preserving local geometric structure rather than summarizing it globally. 
We propose a persistence-based data augmentation framework that encodes local gradient flow regions and their hierarchical evolution using the Morse–Smale complex. 
This representation, compatible with both convolutional and graph neural networks, retains spatially localized topological information across multiple scales. 
Importantly, the augmentation procedure itself is efficient, with computational complexity $O(n \log n)$, making it practical for large datasets. 
We evaluate our method on histopathology image classification and 3D porous material regression, where it consistently outperforms baselines and global TDA descriptors such as persistence images and landscapes. 
We also show that pruning the base level of the hierarchy reduces memory usage while maintaining competitive performance. 
These results highlight the potential of local, structured topological augmentation for scalable and interpretable learning across data modalities.
\end{abstract}

\section{Introduction}
Topological Data Analysis (TDA) is an area of applied mathematics that combines tools from algebraic topology and computer science to describe the shape of data in succinct, informative summaries.
It has achieved prominent success in a wide range of scientific domains, including neuroscience~\cite{giusti2015clique, reimann2017cliques}, materials science~\cite{hiraoka2016hierarchical}, sensor networks~\cite{desilva2007coverage}, and molecular biology~\cite{xia2014persistent}.

In recent years, TDA summaries have also proven useful as input to machine learning (ML) models~\cite{hofer2017deep, carriere2020perslay, Krishnapriyan2020, Hensel2021Survey}.
The features extracted by TDA complement the features learned by the models and thus improve their performance.
Such integration typically relies on stable, vectorized representations of persistence diagrams.
In particular, the introduction of the \emph{persistence image}~\cite{adams2017persistence} and \emph{persistence landscape}~\cite{bubenik2015statistical}---two popular methods for vectorizing persistence diagrams---has spurred advances in domains such as shape classification~\cite{carriere2015stable} and graph learning~\cite{carriere2020perslay}. 
These representations offer compact, differentiable, and interpretable features that can be appended to standard learning pipelines.

On the other hand, because they are derived from persistence diagrams, all such vectorization schemes contain only coarse information: each feature is represented by two values (its birth and death). Meanwhile, the underlying topological features in the domain describe a great deal more about how the shape evolves over the lifetime of the feature.

\paragraph{Our contribution.}
While persistence diagrams are among the most widely used descriptors in TDA, they provide only a global summary of topological features and discard local spatial structure. 
In contrast, Morse–Smale (MS) complexes decompose a domain into regions of uniform gradient flow that retain both topological and geometric locality. 
Our main contribution is an efficient framework for incorporating MS-based topological information into machine learning models. 

We focus on two types of neural networks: Convolutional Neural Networks (CNNs) and Graph Neural Networks (GNNs). 
Despite their effectiveness, they do not inherently take advantage of the topological structure in the data, simply because they are not aware of it. 
To address this limitation, we introduce a novel topological data augmentation framework that enhances neural network performance on both image and graph data. 
Given a grayscale image or a graph with scalar functions on its vertices, we compute its associated Morse--Smale complex. 
From this, we construct a dual representation that encodes the persistence information. 
This yields a hierarchical topological simplification where components with lifetimes below a threshold $\ee$ are iteratively removed. 
The resulting persistence-augmented representations can be stacked as multi-channel images for input to CNNs, or organized into hierarchical graphs for GNNs.

We evaluate our method on two tasks: histopathology image classification and porous material regression. 
Our approach consistently improves performance across both modalities and network types, highlighting the utility of topological priors in deep learning.

\paragraph{Related Work.}
The Morse--Smale complex is a classical construction. In computational topology, it has been successfully applied across a broad range of scientific and engineering domains. 
Its effectiveness in summarizing the topological structure of scalar fields has been demonstrated in applications such as molecular modeling~\cite{edelsbrunner2004morse}, materials science~\cite{hiraoka2016hierarchical, cai2023Rock},
shape segmentation~\cite{Hu2021Segmentation, DeFloriani2015Seg}, and visualization~\cite{gyulassy2007topology, Gyulassy2012Vis}.
These works highlight the MS complex's utility in capturing essential geometric and functional structures in complex data.

Recent years have seen a wealth of work on incorporating topological information into machine learning pipelines. 
Persistence diagrams, persistence images and landscapes have been used to encode global topological summaries into fixed-length vectors suitable for processing by neural networks~\cite{hofer2017deep, carriere2020perslay}. 
In addition, some works have explored incorporating structural topological priors through learned topological loss functions~\cite{moor2020autoencoder}, or by designing TDA-informed models~\cite{Perea2022, Rieck19a}.
However, these approaches primarily emphasize global summaries and do not preserve spatial or local topological structure.

Preserving local topological information remains an open challenge in TDA-ML integration. 
A few notable works address this by studying topology-guided optimization and thus relating the coarse topological features in persistence diagrams to the specific subsets of the data that produce them~\cite{gabrielsson2020topology, poulenard2018topological, Nigmetov2024bigstep, carriere2024diffeo}. These are driven by specific tasks, such as shape correspondence or point cloud classification, formulated as losses on persistence diagrams. 
Our approach is agnostic to the task and constructs a hierarchical dual graph that retains local topological regions and their adjacency across simplification levels.

In contrast to prior work that relies on global summaries or custom architectures, we introduce a general-purpose persistence-augmented pipeline that supports both image and graph data. 
Our method can be integrated with standard convolutional and graph neural networks, enabling local topological information to be leveraged in classification and regression tasks across multiple domains.

\paragraph{Outline.}
The remainder of this paper is organized as follows. 
In Section~\ref{sec:background}, we review relevant background on topological data analysis and persistence representations. 
Section~\ref{sec:methodology} introduces our persistence-based data augmentation pipeline and describes how it integrates with standard neural network architectures. 
In Section~\ref{sec:experiments}, we evaluate our method on two datasets---one for image classification and one for 3D regression---using both convolutional and graph neural networks. 
Section~\ref{sec:discussion} discusses the results, limitations, and future directions, and we conclude in Section~\ref{sec:conclusion}.

\section{Background}
\label{sec:background}
In this section, we briefly review Morse theory for smooth scalar functions and its discrete counterpart, discrete Morse theory. For full details, we refer the reader to Milnor~\cite{milnor1963morse} for the smooth setting and Forman~\cite{forman1998morse} for the discrete case.

\subsection{Morse Function and the Morse--Smale Complex}
Let $f: M \to \mathbb{R}$ be a smooth real-valued function defined on a compact $d$-dimensional manifold $M$. 
The function $f$ is a \emph{Morse function} if all of its critical points are nondegenerate, meaning the Hessian matrix at each critical point is non-singular, and all critical values are distinct.
The \emph{index} of a critical point is the number of negative eigenvalues of the Hessian.
An \emph{integral line} of $f$ is a maximal path within $M$ whose tangent vectors are aligned with the gradient of $f$. 
These paths begin and end at critical points where the gradient vanishes.
\emph{Ascending} and \emph{descending manifolds} are defined by grouping integral lines that originate from or terminate at the same critical point, respectively.
The \emph{Morse--Smale complex} $\Gamma$ partitions the manifold into regions comprising integral lines with common origin and destination. 
In Morse--Smale functions, such lines connect critical points of differing indices.

A critical point of index $n$ is the origin of an ascending manifold of dimension $d - n$, and likewise, the destination of a descending manifold of dimension $n$. 
These manifolds intersect transversely. For a pair of critical points $a$ and $b$ such that the index of $a$ is one less than that of $b$, the intersection of the ascending manifold of $a$ and the descending manifold of $b$ is either empty or forms a 1-dimensional manifold.
The \emph{nodes} and \emph{arcs} of the Morse--Smale complex correspond to the critical points and these 1-dimensional intersections, respectively. 
Together, they form the one-skeleton---a combinatorial graph structure that captures essential features of $f$. 
The neighborhood $N_a$ of a node $a$ in $\Gamma$ consists of all nodes connected to $a$ via arcs in the complex.

\subsection{Discrete Morse Theory}
Discrete Morse theory~\cite{forman1998morse} generalizes smooth Morse theory to combinatorial settings, enabling critical point analysis on discrete structures such as CW-complexes. 
In this framework, a function defined on a regular CW--complex satisfies specific conditions that allow identifying critical cells and constructing discrete gradient fields, analogous to gradient flows in the smooth case. 
The resulting combinatorial structures, particularly discrete Morse--Smale complexes, capture essential topological features of the domain.

A brief overview is as follows: a discrete Morse function assigns scalar values to cells under constraints that limit how many faces and cofaces can have nonincreasing or nondecreasing values. 
Critical cells are those not paired via these constraints, and discrete vector fields are formed by matching adjacent cells to model flow. 
Paths along these vectors, known as $V$-paths, behave analogously to integral lines in smooth theory.

We use discrete Morse--Smale complexes derived from these vector fields to obtain topological summaries of images and graphs. 
Full definitions, including the formal properties of discrete Morse functions, vector fields, and cancellation operations, are provided in Appendix~\ref{appendix:discrete_morse}.

\subsection{Persistence-based Simplification}
A scalar function $f$ can be simplified by canceling pairs of critical points, smoothing both the gradient vector field and $f$ itself, see~\cite{edelsbrunner2004morse} for details. 
Critical point pairs are prioritized by \emph{persistence}, defined as the absolute difference in $f$-values between the two points, so that low-persistence features are removed first. 
See Appendix~\ref{appendix:topology} for definitions of persistence and persistence diagram.

A cancellation is valid if the two critical points are connected by exactly one arc in the Morse--Smale (MS) complex and their indices differ by one. 
Configurations where multiple arcs connect two points, known as \emph{strangulations} or \emph{pouches}, cannot be canceled through local perturbations.

Let $\Gamma$ denote the MS complex of $f$ on a closed $d$-manifold $ M$. 
Consider an arc $ a$ connecting a lower node $l$ of index $i$ and an upper node $u$ of index $i+1$. 
Let $N_l$ and $N_u$ denote the sets of nodes adjacent to $l$ and $u$, respectively. 
The combinatorial cancellation modifies $\Gamma$ by connecting each critical point of index $i+1$ in $N_l$ to each critical point of index $i$ in $N_u$, and then removing all arcs incident to $l$ and $u$ along with the nodes themselves.

This combinatorial change also induces a geometric modification. 
The descending manifolds of $u$ are merged with the descending manifolds of its neighboring $(i+1)$-index nodes, while the ascending manifolds of $l$ are merged with the ascending manifolds of its neighboring $i$-index nodes. 
After cancellation, the MS complex differs only where new intersections between the updated ascending and descending manifolds occur. 
Although up to $|N_l| \times |N_u|$ new arcs and corresponding cells may be introduced, the total number of critical points decreases by two, and many newly introduced arcs are typically removed during subsequent cancellations, such as in saddle-extremum simplifications.

\section{Methodology}
\label{sec:methodology}

Instead of resolving the boundaries of the cells in the Morse--Smale complex, which is the source of most complications for their algorithmic construction~\cite{edelsbrunner2004morse}, we simply identify the source and destination of the integral lines of every vertex in our domain. For most vertices, these source and destination are minima and maxima. If a vertex falls on an integral line that goes to a saddle, we continue it from the saddle until we reach an extremum, effectively assigning every vertex a unique minimum--maximum pair. This has the effect of perturbing the function infinitesimally, breaking any degeneracies at saddle points.

This perturbation offers several advantages: it allows us to construct the base MS complex directly on discrete domains such as images or graphs, without requiring higher-dimensional cells, and it reduces the complexity of the construction. Specifically, the base MS complex can be built in $O(n)$ time, where $n$ is the number of vertices or pixels, by tracing gradient paths from each vertex. The full $k$-level simplification hierarchy can then be computed in $O(n\log n + kn)$ time: the $O(n \log n)$ term accounts for sorting the persistence pairs by their persistence values, and each of the $k$ simplification levels requires a linear pass to update region labels.

\subsection{Constructing the Morse--Smale Complex}

We begin by constructing the Morse--Smale complex on two types of discrete domains: grayscale images and graphs with scalar functions on their vertices. 
In both cases, the domain is modeled as a discrete scalar field, and the MS complex captures the topological structure induced by the gradient of the scalar function.

\paragraph{Image data.}
Let $I : \Omega \to \mathbb{R}$ be a grayscale image, where $\Omega \subset \mathbb{Z}^2$ denotes the pixel grid and $I(p)$ is the intensity value at pixel $p \in \Omega$. 
Each pixel is treated as a vertex in a regular grid graph, and we use 4-connectivity to define adjacency: two pixels $p$ and $q$ are neighbors if $\|p - q\|_1 = 1$, i.e., they share a horizontal or vertical edge.

The intensity values $I(p)$ define a discrete scalar function on this graph. 
Using discrete Morse theory, we construct a discrete gradient vector field by pairing each vertex with a neighbor of higher intensity, when possible. 
Vertices that are unmatched under this rule are designated as critical points---corresponding to local minima or maxima depending on their local configuration.

We trace discrete gradient paths (or integral lines) by following these pairings. 
Pixels that share a common origin (minimum) or destination (maximum) under these paths are grouped into ascending and descending manifolds, respectively.
Because all points on a gradient path share their origin and destination, all such labels can be computed in linear time. 
The intersections of these manifolds partition the image into cells that define the MS complex.

\paragraph{Graph data.}
Let $G = (V, E)$ be an undirected graph, and let $f : V \to \mathbb{R}$ assign a scalar value to each vertex. 
We interpret this as a discrete scalar field over the graph domain. 
Two vertices $u, v \in V$ are adjacent if $(u, v) \in E$.

The MS complex is constructed analogously: edges are directed from lower to higher function value to simulate gradient flow. 
Critical vertices are those not matched via the gradient vector field. 
By tracing discrete gradient paths along the edges of the graph, we obtain a segmentation of the vertex set into ascending and descending regions centered at critical points. 
Their intersections define the cells of the Morse--Smale complex on the graph.

This construction encodes the topological structure of the input data, capturing how regions of the domain are organized by the flow of the underlying scalar function.
It can be computed in linear time, in contrast to the common perception of topological data analysis as computationally intensive.

\subsection{Dual Representation as a Topological Graph}

To obtain a compressed and structured representation of this topological information, we construct a dual graph $\mathcal{G} = (\mathcal{V}, \mathcal{E})$.

Each vertex $v \in \mathcal{V}$ corresponds to a 2-cell $R_{(m, M)} \subseteq \Omega$ in the MS complex, where all points in $R_{(m, M)}$ flow from a common local minimum $m$ to a common local maximum $M$ under the discrete gradient field. That is,
\[
\mathcal{V} = \left\{ v_{(m, M)} \,\middle|\, R_{(m, M)} = \{ x \in \Omega \mid \phi(x) = (m, M) \} \right\},
\]
where $\phi(x)$ denotes the ordered pair of critical points reached by following the gradient path through $x$.

Edges in $\mathcal{E}$ are defined between regions that are adjacent in the MS complex. 
Two nodes $v_{(m_1, M_1)}$ and $v_{(m_2, M_2)}$ are connected by an edge if there exists a shared boundary between their corresponding regions in the domain. 
Specifically, $(v_1,v_2)$ iff $\exists x_1 \in v_1, x_2 \in v_2$ and $(x_1,x_2) \in E$.
This dual graph encapsulates the flow-based partitioning of the domain in a condensed, combinatorial form. 
We define a persistence weight for each edge based on the birth-death pair corresponding to the saddle point that lies between the two adjacent regions.
Specifically, let $(c_i, c_j)$ be a persistence pair of index $(i, i+1)$ whose cancellation would merge $R_{(m_1, M_1)}$ and $R_{(m_2, M_2)}$; then we define:
\[
w\left( v_{(m_1, M_1)}, v_{(m_2, M_2)} \right) = |f(c_i) - f(c_j)|.
\]

\begin{figure}
    \centering
    \includegraphics[width=0.8\linewidth]{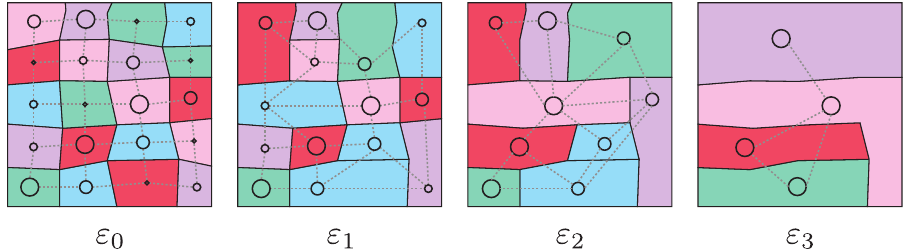}
    \caption{Dual graph and simplification}
    \label{fig:dualGraph}
\end{figure}

The resulting weighted graph $\mathcal{G}$ provides a topologically-informed abstraction of the original data. 
It reduces redundancy by grouping all vertices (pixels or original graph nodes) that exhibit identical topological behavior under $f$, and captures adjacency and persistence structure in the edge set. 
This dual representation forms the basis for hierarchical simplification and serves as input to graph-based learning models.
See Figure~\ref{fig:dualGraph} for an example.

\subsection{Hierarchical Simplification via Persistence}

Topological simplification is performed by canceling critical point pairs in the MS complex based on their persistence. 
For a pair $(c_i, c_j)$ consisting of a saddle and an extremum, persistence is defined as the absolute difference $|f(c_i) - f(c_j)|$. 
Lower-persistence pairs typically correspond to topological noise, while higher-persistence features represent more prominent structures.

To build a hierarchy, we define a sequence of increasing thresholds $\ee_1 < \ee_2 < \cdots < \ee_k$, and iteratively cancel all persistence pairs with $\text{pers}(c_i, c_j) < \ee_\ell$ at each level $\ell$. 
This yields a series of simplified MS complexes $\Gamma^{(\ee_1)}, \Gamma^{(\ee_2)}, \dots, \Gamma^{(\ee_k)}$, and corresponding dual graphs $\mathcal{G}^{(\ee_1)}, \mathcal{G}^{(\ee_2)}, \dots$. 

Each graph $\mathcal{G}^{(\ee_\ell)} = (\mathcal{V}_\ell, \mathcal{E}_\ell)$ encodes the MS regions as nodes and their adjacency as edges. 
In addition to intra-level structure, we retain inter-level edges that track the merging of regions across simplification steps. 
Specifically, if a region $R \in \mathcal{V}_\ell$ is merged into $R' \in \mathcal{V}_{\ell+1}$, we add a directed edge from $R$ to $R'$. 
This defines a hierarchical graph structure with cross-scale connectivity:
\[
\mathcal{H} = \left( \bigcup_{\ell=1}^k \mathcal{V}_\ell,\, \bigcup_{\ell=1}^k \mathcal{E}_\ell \cup \mathcal{E}_{\ell \to \ell+1} \right).
\]

\begin{figure}[ht]
    \centering
    \includegraphics[width=0.65\linewidth]{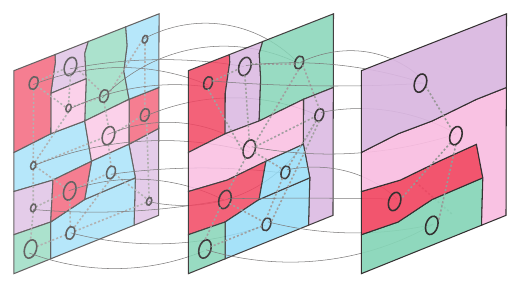}
    \caption{
    Construction of the hierarchical graph input for GNNs. Each layer represents a simplification level \( \mathcal{V}_\ell \), where nodes correspond to Morse–Smale regions defined by \((m, M)\) pairs. Intra-level edges (dashed lines) connect adjacent regions, while inter-level edges (solid lines) indicate merge relationships as simplification progresses.
    }
    \label{fig:gnnInput}
\end{figure}

This hierarchy supports two representations for learning:
\begin{itemize}
    \item For \textbf{CNNs}, we discard the inter-level edges and encode the region structure of each simplification level into separate image channels. 
    Each channel corresponds to the partition of the domain into regions in $\mathcal{V}_\ell$, where each pixel is assigned a label identifying its associated minimum--maximum pair $(m, M)$. 
    This representation retains local minima and maxima at different scales but does not preserve explicit topological transitions between levels. Instead they are encoded implicitly by the labels in the channels of individual pixels.
    \item For \textbf{GNNs}, we retain the full hierarchical graph $\mathcal{H}$, where each node represents a region from a given simplification level $\mathcal{V}_\ell$, and edges include both intra-level connections (capturing spatial adjacency) and inter-level edges (capturing merge relationships across levels). This structure supports message passing both within and between scales, allowing the model to learn how local topological features evolve through the simplification hierarchy. See Figure~\ref{fig:gnnInput} for an illustration.
\end{itemize}

\subsection{Neural Network Architectures}

To evaluate the effectiveness of persistence-augmented representations, we employ standard baseline models for both image and graph data. 
Our focus is on isolating the contribution of the augmented topological features, so we intentionally use minimal neural architectures without architectural tuning or task-specific modifications.

For image data, we use a conventional convolutional neural network (CNN) consisting of two convolutional layers with ReLU activations, followed by max pooling and two fully connected layers. 
Persistence-augmented inputs are provided as multi-channel images, with each channel corresponding to a different level of topological simplification. 
All image inputs are processed with the same CNN architecture.
For graph data, we adopt a basic graph neural network (GNN) designed for graph-level prediction. 
Each input graph represents the dual MS complex, where nodes correspond to topological regions and edges encode both intra-level adjacency and inter-level hierarchical merges. 
The GNN consists of two message passing layers, followed by global pooling to aggregate node features into a graph-level representation, and a fully connected layer to produce the final prediction.
Both CNN and GNN models are used for classification and regression tasks, depending on the dataset. Specific task setups are detailed in the experimental sections.

\section{Experiments}
\label{sec:experiments}
We evaluate the effectiveness of persistence-based topological augmentation on both image and graph data by comparing classification and regression performance using standard CNN and GNN architectures. 
Our goal is to isolate the contribution of topological information in learned representations, rather than optimize architectural complexity. 
All experiments were conducted using consistent training setups, including five-fold cross-validation, standardized data preprocessing, and model selection via grid search over learning rate and weight decay parameters.

\subsection{Histopathology Image Classification}

\begin{figure}
    \centering
    \includegraphics[width=0.7\linewidth]{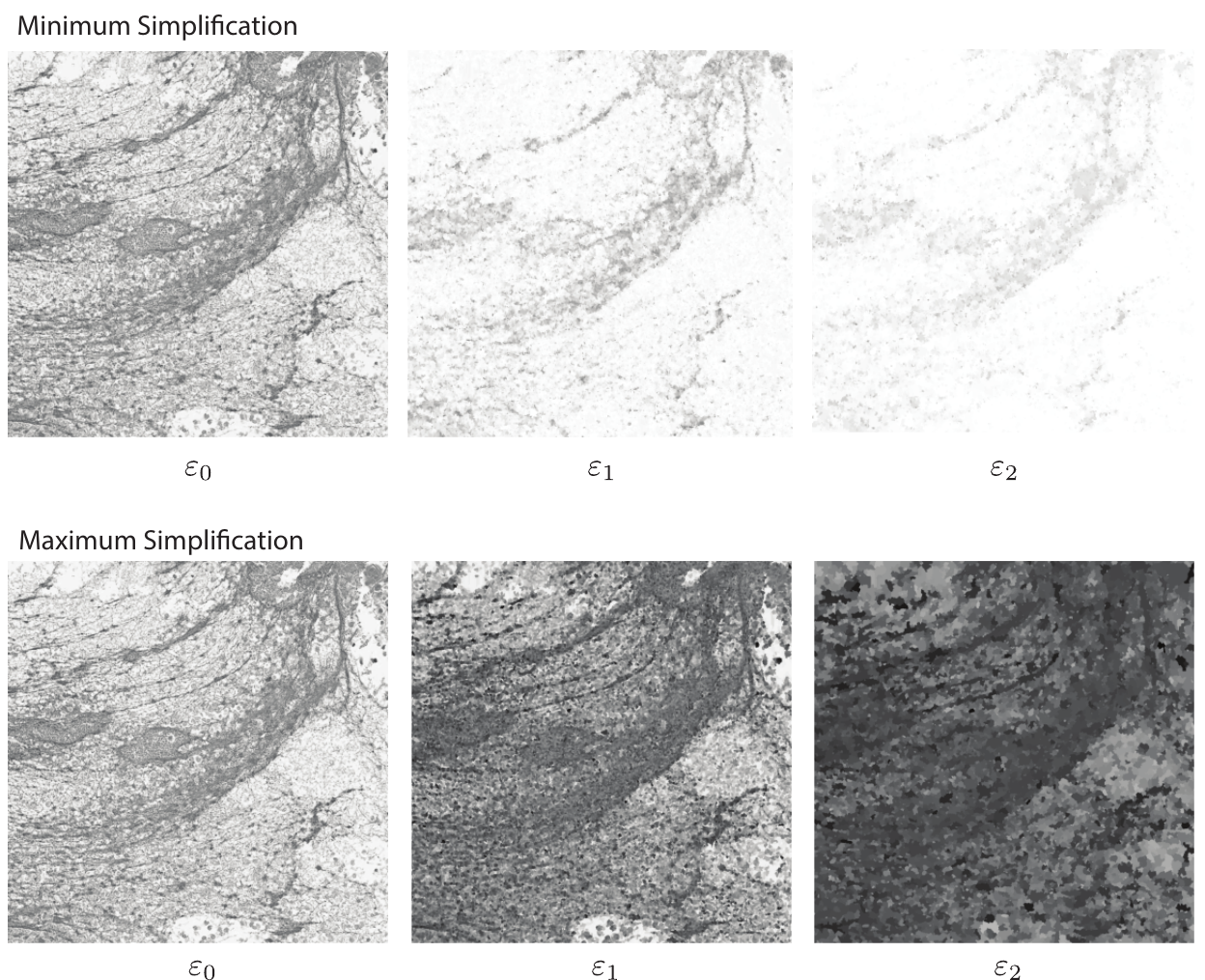}
    \caption{Visualization of different levels of simplification}
    \label{fig:cancerExample}
\end{figure}

The first dataset consists of 1,144 RGB histopathology images, each of resolution $1024 \times 1024$, categorized into three classes: non-tumorous (47\%), necrotic tumor (23\%), and viable tumor (30\%)~\cite{leavey2019osteosarcoma}.
We convert all images to grayscale and treat them as discrete scalar fields. 
From each image, we compute a MS complex using 4-connectivity and construct a corresponding dual representation. 
To generate hierarchical representations, we define four levels of simplification based on persistence thresholds: the original unsimplified image (level 0), removal of low-persistence features representing approximately 30\% and 65\% of components (levels 1 and 2), and an extreme simplification retaining only the global minimum and maximum (level 3).
See Figure~\ref{fig:cancerExample} for a visualization of these levels of simplification with respect to minimum and maximum.

These simplification levels serve as input for two neural architectures. 
For CNNs, the scalar values of each simplification level are encoded as separate image channels, yielding a multi-channel input. 
For GNNs, we construct a dual graph at each level and connect consecutive levels through inter-layer edges that record which regions were merged during simplification.

In addition to this persistence-augmented pipeline, we evaluate two topological descriptors commonly used in prior TDA literature: the \emph{persistence image} and the \emph{persistence landscape}. 
Both summarize a persistence diagram as a vectorized feature map that can be appended to standard model inputs. The persistence image discretizes the birth-death plane into a fixed-resolution grid, applying a Gaussian kernel centered on each persistence point. 
The persistence landscape encodes a collection of piecewise-linear functions representing feature lifetimes across scales. 
A detailed description of both constructions is provided in Appendix~\ref{appendix:tda_representations}.

These representations provide compact global summaries of the topological structure but discard spatial localization and adjacency relationships. 
In contrast, our persistence-augmented input preserves local topological information through region-based grouping and hierarchical adjacency across simplification levels. 
This enables our models to learn not only what features are persistent, but also where and how they interact within the data domain.

We evaluate six model configurations: (1) CNN on original grayscale images, (2) CNN with multi-channel persistence augmentation, (3) CNN with appended persistence image, (4) CNN with appended persistence landscape, (5) GNN trained on the full hierarchical dual graph, and (6) GNN trained on the same graph with the base level removed. 
All models are trained using cross-entropy loss, and performance is measured using accuracy and Cohen's kappa.

\subsection{3D Porous Material Regression}

The second dataset consists of 1,700 volumetric samples generated via the Puma and Palabos simulation tools. Each sample is a $256^3$ binary volume, where 0 represents air and 1 represents solid material. 
The dataset includes associated porosity values and permeability estimates, with higher porosity typically correlating with higher permeability. 
Each sample is labeled with a single target value. 

We compute a distance transform for each voxel to its nearest obstacle cell and treat this distance field as the input scalar function for persistence computation. 
We then extract the MS complex and construct the corresponding dual graph. As in the image domain, we generate four simplification levels using increasing persistence thresholds and encode the results as either multi-channel volumetric inputs (for CNNs) or hierarchical graphs (for GNNs).

To increase task difficulty and better assess the quality of the topological signal, we restrict our evaluation to samples with fixed porosity levels of 0.6, 0.7, and 0.75. 
When the task is performed on 735 sampled data points across the full range of porosity values using the original voxel input, we achieve a high test $R^2$ score of 0.96, and a further improvement to 0.99 with persistence-augmented input. 
Therefore, we focus on the more challenging and restricted setting of single-porosity subsets. 
For each porosity level, we randomly sample 245 data points from the available samples at that porosity and train separate models using mean squared error loss. 
Performance is evaluated using mean and standard deviation of test $R^2$ scores across five-fold cross-validation.

\subsection{Results}

Table~\ref{tab:histopath_results} presents classification results on the histopathology dataset. 
CNNs trained with persistence-augmented multi-channel inputs significantly outperform the baseline, and further gains are observed when using GNNs trained on the full hierarchical dual graph. 
While persistence images and landscapes yield modest improvements over the original input, their performance remains consistently lower than that of our structured topological augmentation.

Regression results for the porous material dataset are shown in Table~\ref{tab:porous_results}. 
Across all porosity levels, persistence-augmented inputs improve model generalization compared to the baseline, with particularly large gains in the more challenging settings at porosity levels 0.6 and 0.7. 
GNNs that incorporate hierarchical structure achieve the highest performance overall, reinforcing the benefit of preserving topological relationships across simplification levels.

\begin{table}[ht]
\centering
\caption{Classification performance on the histopathology dataset. Results are reported as mean accuracy / Cohen's kappa with standard deviation over 5-fold cross-validation.}
\label{tab:histopath_results}
\begin{tabular}{lcc}
\toprule
\textbf{Model} & \textbf{Accuracy (\%)} & \textbf{Cohen's Kappa (\%)} \\
\midrule
CNN (Original) & 84.0 \(\pm\) 0.7 & 71.0 \(\pm\) 0.6 \\
CNN + Persistence Image & 88.5 \(\pm\) 0.5 & 80.2 \(\pm\) 0.8 \\
CNN + Persistence Landscape & 89.1 \(\pm\) 0.6 & 82.3 \(\pm\) 0.9 \\
CNN + Persistence Augmentation & \textbf{95.0} \(\pm\) 0.5 & \textbf{91.0} \(\pm\) 0.6 \\
GNN (Full Hierarchy) & 92.0 \(\pm\) 1.1 & 89.0 \(\pm\) 1.0 \\
GNN (w/o Base Layer) & 90.5 \(\pm\) 1.0 & 86.5 \(\pm\) 1.3 \\
\bottomrule
\end{tabular}
\end{table}

\begin{table}[ht]
\centering
\caption{Regression performance on the porous material dataset. Results are reported as mean test \(R^2\) scores \(\pm\) standard deviation over 5-fold cross-validation.}
\label{tab:porous_results}
\begin{tabular}{lccc}
\toprule
\textbf{Model} & \textbf{Porosity 0.6} & \textbf{Porosity 0.7} & \textbf{Porosity 0.75} \\
\midrule
CNN (Original) & 0.74 \(\pm\) 0.04 & 0.28 \(\pm\) 0.06 & 0.88 \(\pm\) 0.02 \\
CNN + Persistence Image & 0.80 \(\pm\) 0.03 & 0.53 \(\pm\) 0.05 & 0.90 \(\pm\) 0.01 \\
CNN + Persistence Landscape & 0.81 \(\pm\) 0.03 & 0.56 \(\pm\) 0.04 & 0.91 \(\pm\) 0.01 \\
CNN + Persistence Augmentation & \textbf{0.88} \(\pm\) 0.02 & 0.72 \(\pm\) 0.03 & \textbf{0.95} \(\pm\) 0.01 \\
GNN (Full Hierarchy) & 0.86 \(\pm\) 0.03 & \textbf{0.76} \(\pm\) 0.04 & 0.92 \(\pm\) 0.01 \\
GNN (w/o Base Layer) & 0.84 \(\pm\) 0.03 & 0.70 \(\pm\) 0.04 & 0.91 \(\pm\) 0.01 \\
\bottomrule
\end{tabular}
\end{table}

\section{Discussion}
\label{sec:discussion}
Our experiments demonstrate that persistence-based topological augmentation substantially improves performance across both classification and regression tasks, outperforming models trained on original data as well as those enhanced with global persistence summaries such as persistence images and landscapes. 
These results validate the utility of fine-grained topological information, particularly when encoded through region-based representations and hierarchical simplification.

Unlike prior work relying on global persistence features, our framework is compatible with both convolutional and graph-based architectures. 
This flexibility enables application to a wider variety of data types, including domains where graph structures arise naturally or must be constructed from spatial adjacency. 
Notably, persistence image and landscape descriptors are fundamentally limited to vector representations and are therefore not easily applicable in graph-based learning frameworks. 
In contrast, our method enables learning over rich, structured topological and geometric signals using GNNs.
Regression results also show that GNNs outperform CNNs primarily at porosity level 0.7. 
While the performance gap is more modest at other porosity levels, it is important to note that we have used simple, baseline architectures for both CNNs and GNNs to ensure fair comparisons. 
We did not explore architectural tuning or deeper models, particularly for the GNN setting. 
With more sophisticated designs, such as attention mechanisms, residual connections, or advanced pooling strategies, we expect GNNs could better exploit the hierarchical graph structure and achieve further improvements.

We note that the persistence-augmented CNN receives a multi-channel input encoding multiple simplification levels, which provides the network with strictly more information than the single-channel baseline. 
Part of the performance gain may therefore reflect the richer input representation rather than the topological signal alone. 
However, the fact that our structured augmentation substantially outperforms persistence images and landscapes---which also augment the baseline with additional topological information, albeit in a different form---suggests that the spatially localized, hierarchical encoding is a key factor in the improvement.

One practical challenge of our approach lies in the increased cost of training with augmented data. 
The augmented inputs lead to increased memory usage and longer training times, particularly in the GNN setting where message passing must account for hierarchical interconnections. 
Optimizing training efficiency, for example, through data compression, batching strategies, or sparsity-aware architectures, presents an important direction for future work.

Additionally, while our current pipeline uses fixed simplification thresholds across all datasets, this uniform strategy may not capture dataset-specific variations in scale or complexity. 
A promising extension would be to make the simplification process differentiable and data-adaptive, allowing the model to learn which topological features are most relevant during training.

Finally, our current GNN pipeline focuses on graph-level prediction tasks, but the structured nature of the hierarchical graph also lends itself to finer-grained analysis. 
Given appropriate supervision, this framework could be extended to node-level or edge-level tasks such as region labeling, boundary detection, or spatial segmentation, which would enable more localized topological inference and broader application to scientific and biomedical domains.

While removing the base level from the hierarchical GNN representation results in a modest drop in predictive performance, we observe a corresponding reduction in memory usage and computational cost. 
For large-scale datasets or applications where resources are limited, this trade-off may be worthwhile, particularly when top-level structural features dominate the task. 
This suggests that the granularity of the hierarchy can be tuned to balance performance and efficiency depending on application needs.

Overall, this work highlights the potential of local, structured topological representations to enhance neural learning across modalities. 
Our results suggest that topology-aware representations can offer a generalizable approach to improving model performance on complex scientific and geometric data.

\section{Conclusion}
\label{sec:conclusion}
We introduced a persistence-based data augmentation framework that integrates topological summaries with standard neural network architectures for both image and graph data. 
By leveraging the Morse--Smale complex and its hierarchical simplification, our method encodes local topological structure in a form compatible with convolutional and graph neural networks. 
Through experiments on both classification and regression tasks, we demonstrated that this augmentation strategy consistently improves performance over original inputs and global topological descriptors such as persistence images and landscapes.

Our framework offers a general and scalable approach to incorporating topological priors in deep learning, without requiring task-specific architectures or losses. 
Future work will focus on optimizing computational efficiency, extending the method to node- and edge-level prediction tasks, and exploring learnable topological simplification as part of the training pipeline. 
Overall, our results highlight the value of local topological information for improving model performance and interpretability across domains.

\paragraph*{Acknowledgements.} This work was supported by the U.S.\ Department of Energy, Office of Science, Office of Advanced Scientific Computing Research under Contract Number DE-AC02-05CH11231 at Lawrence Berkeley National Laboratory. This research used the Lawrencium cluster at Lawrence Berkeley National Laboratory.

\bibliographystyle{unsrt}
\bibliography{PersNeuralNet}

\appendix
\section{Discrete Morse Theory Formal Definitions}
\label{appendix:discrete_morse}

We summarize the main notions of discrete Morse theory following Forman~\cite{forman1998morse}. 

A \emph{$d$-cell} is a topological space homeomorphic to a $d$-dimensional closed ball $B^d = \{ x \in \mathbb{R}^d : \|x\| \leq 1 \}$. 
Given two cells $\sigma$ and $\tau$, we write $\sigma < \tau$ to indicate that $\sigma$ is a face of $\tau$, meaning the vertices of $\sigma$ form a proper subset of those of $\tau$. 
If $\dim(\sigma) = \dim(\tau) - 1$, then $\sigma$ is called a \emph{facet} of $\tau$, and $\tau$ a \emph{cofacet} of $\sigma$.

A \emph{finite CW-complex} $X$ is a topological space built from a finite sequence
\[
\emptyset = X_{-1} \subset X_0 \subset X_1 \subset \cdots \subset X_n = X,
\]
where $X_i$ is formed by attaching $i$-cells to $X_{i-1}$. 
A \emph{regular} CW-complex imposes additional structure: for any two incident cells $\rho$ and $\tau$ with $\dim(\tau) = \dim(\rho) - 2$, there exist exactly two distinct intermediate cells $\sigma_1$ and $\sigma_2$ such that $\tau < \sigma_j < \rho$ for $j = 1,2$. 
This ensures well-behaved attachments, forcing boundaries of cells to be homeomorphic to spheres of appropriate dimension.

Let $K$ denote a regular CW-complex approximating the manifold $M$. 
A function $f: K \to \mathbb{R}$ assigning real values to each cell of $K$ is called a \emph{discrete Morse function} if for every $d$-cell $\sigma \in K$,
\[
|\{ \tau^{(d+1)} > \sigma \mid f(\tau) \leq f(\sigma) \}| \leq 1 \quad \text{and} \quad |\{ \gamma^{(d-1)} < \sigma \mid f(\gamma) \geq f(\sigma) \}| \leq 1.
\]
A cell $\sigma$ is \emph{critical} if both counts are zero: no lower-dimensional faces have greater function values and no higher-dimensional cofaces have smaller function values.

In the discrete setting, a \emph{vector} is an ordered pair $(\sigma^{(d)}, \tau^{(d+1)})$ where $\sigma$ is a facet of $\tau$, intuitively indicating a direction of flow from $\sigma$ to $\tau$. 
A \emph{discrete vector field} $V$ is a collection of such pairs where each cell appears in at most one pair.

Given a discrete vector field $V$ on $K$, a \emph{$V$-path} is a sequence
\[
\sigma_0^{(d)}, \tau_0^{(d+1)}, \sigma_1^{(d)}, \tau_1^{(d+1)}, \ldots, \tau_r^{(d+1)}, \sigma_{r+1}^{(d)}
\]
such that for each $i$, $(\sigma_i, \tau_i) \in V$ and $\tau_i > \sigma_{i+1} \neq \sigma_i$. 
A $V$-path represents the discrete analogue of an integral curve in a smooth vector field.

A discrete vector field is called a \emph{discrete gradient field} if it contains no nontrivial closed $V$-paths (i.e., no cycles). 
In our framework, we use $V$-paths to construct the Morse--Smale complex associated with a discrete gradient field.

\section{Topological Notions}
\label{appendix:topology}

Persistent homology is a method from computational topology that captures the multiscale topological structure of data. 
It does so by analyzing how homological features (e.g., connected components, loops, voids) appear and disappear across a filtration of simplicial complexes.

\subsection{Simplicial Complexes and Filtrations}

Let $X$ be a finite metric space or a scalar function $f: X \to \mathbb{R}$. 
A \emph{simplicial complex} $K$ is a finite set of simplices (vertices, edges, triangles, etc.) that is closed under inclusion of faces. Given a function $f$, we define a \emph{sublevel set filtration} by taking
\[
K_t := \{ \sigma \in K \mid \max_{v \in \sigma} f(v) \leq t \}, \quad t \in \mathbb{R}.
\]
This produces a nested sequence of simplicial complexes
\[
K_{t_0} \subseteq K_{t_1} \subseteq \cdots \subseteq K_{t_n},
\]
called a \emph{filtration}, where $t_0 < t_1 < \cdots < t_n$ are critical values of $f$.

Alternatively, in the metric setting, one may use a Vietoris–Rips or Čech filtration. 
For a point cloud $X \subset \mathbb{R}^d$, the Vietoris–Rips complex $\text{VR}_\epsilon(X)$ is defined as:
\[
\text{VR}_\epsilon(X) := \{ \sigma \subseteq X \mid \text{diam}(\sigma) \leq \epsilon \},
\]
and we build a filtration $\text{VR}_{\epsilon_0}(X) \subseteq \text{VR}_{\epsilon_1}(X) \subseteq \cdots$ as $\epsilon$ increases.

\subsection{Homology and Persistence Modules}

For a fixed dimension $k \geq 0$, let $H_k(K_t; \mathbb{F})$ denote the $k$-th simplicial homology group with coefficients in a field $\mathbb{F}$ (e.g., $\mathbb{F} = \mathbb{Z}_2$). 
The inclusion maps $K_{t_i} \hookrightarrow K_{t_j}$ for $t_i \leq t_j$ induce homomorphisms between homology groups:
\[
\phi_{i,j}^k: H_k(K_{t_i}; \mathbb{F}) \to H_k(K_{t_j}; \mathbb{F}).
\]
The sequence $\{ H_k(K_t; \mathbb{F}) \}$ together with the maps $\phi_{i,j}^k$ forms a \emph{persistence module}. 
The structure theorem for persistence modules over a field (when the filtration is finite) states that such a module decomposes uniquely into interval modules:
\[
H_k(K_t) \cong \bigoplus_{i} \mathbb{I}_{[b_i, d_i)},
\]
where each interval \([b_i, d_i)\) represents a \(k\)-dimensional homological feature that appears (is "born") at time $b_i$ and disappears (is "killed") at time $d_i$.

\subsection{Persistence Diagrams}

The multiset $\mathcal{D}_k = \{ (b_i, d_i) \} \subset \mathbb{R}^2$ of birth-death pairs corresponding to this decomposition is called the \emph{persistence diagram} in dimension $k$. 
It encodes the lifetime of each topological feature across the filtration. 
Points near the diagonal represent short-lived (low-persistence) features, often considered noise, while points far from the diagonal correspond to prominent topological structures.

Formally, persistence diagrams lie in the space of finite multisets of $\Delta^+ := \{ (x, y) \in \mathbb{R}^2 \mid x < y \} \cup \{ \text{diagonal} \}$, where the diagonal accounts for zero-persistence features.

\subsection{Stability and Distance Metrics}

Persistence diagrams are stable under perturbations of the input. 
For example, given two filtrations $K$ and $K'$ with corresponding persistence diagrams $\mathcal{D}, \mathcal{D}'$, the \emph{bottleneck distance} is defined as:
\[
d_B(\mathcal{D}, \mathcal{D}') := \inf_{\gamma} \sup_{x \in \mathcal{D}} \| x - \gamma(x) \|_\infty,
\]
where $\gamma$ ranges over all bijections between $\mathcal{D}$ and $\mathcal{D}' \cup \Delta$. 
The bottleneck and Wasserstein distances are commonly used to compare persistence diagrams and are known to satisfy stability theorems under tame filtrations~\cite{cohen2007stability}.

\section{Persistence Diagram Representations}
\label{appendix:tda_representations}

Persistence diagrams are commonly used to summarize the birth and death of topological features in data. To integrate these summaries into standard machine learning pipelines, various vectorization techniques have been proposed. 
In our experiments, we compare two such descriptors: the \emph{persistence image} and the \emph{persistence landscape}.

\subsection{Persistence Images}

A persistence diagram $\mathcal{D} = \{ (b_i, d_i) \}_{i=1}^n$ consists of birth-death pairs where $b_i$ and $d_i$ denote the birth and death times of topological features. 
To construct a persistence image~\cite{adams2017persistence}, we first map each point to the \emph{birth-persistence} plane:
\[
(b_i, p_i) \quad \text{where} \quad p_i = d_i - b_i.
\]
Each point is then convolved with a Gaussian kernel:
\[
\rho_i(x, y) = \exp\left(-\frac{(x - b_i)^2 + (y - p_i)^2}{2\sigma^2}\right),
\]
and the resulting functions are summed to form a continuous surface. 
This surface is discretized over a grid of fixed resolution, yielding a persistence image $P \in \mathbb{R}^{m \times m}$. 
In practice, the image is flattened and concatenated to the model input or appended to the feature representation.

Persistence images are stable with respect to small perturbations of the diagram and allow for direct integration into neural networks as fixed-length vectors.

\subsection{Persistence Landscapes}

The persistence landscape~\cite{bubenik2015statistical} transforms a persistence diagram into a sequence of piecewise-linear functions. For each birth-death pair $(b_i, d_i)$, we define a tent-shaped function:
\[
\lambda_i(t) = \max\{ 0, \min(t - b_i, d_i - t) \},
\]
which peaks at $(b_i + d_i)/2$ with height $(d_i - b_i)/2$. The $k$-th landscape function $\lambda_k(t)$ is then defined pointwise as the $k$-th largest value among all $\lambda_i(t)$ at each $t$:
\[
\lambda_k(t) = \text{$k$-th largest of } \{ \lambda_1(t), \lambda_2(t), \dots, \lambda_n(t) \}.
\]

In practice, a finite number $K$ of landscape layers are retained and discretized over a uniform grid of $T$ evaluation points. This results in a matrix $\Lambda \in \mathbb{R}^{K \times T}$, which is flattened and appended to the model input.

Persistence landscapes preserve stability and provide interpretable information about feature prominence across scales. 
They also permit functional operations such as integration and averaging.

\subsection{Implementation Details}

For both persistence images and landscapes, we use the \texttt{Giotto-TDA} Python library. 
Persistence diagrams are computed from the sublevel set filtration of the grayscale image or distance-transformed 3D volume. 
For persistence images, we use a $20 \times 20$ grid and $\sigma = 0.1$; for landscapes, we retain the top $K=5$ landscape layers and evaluate over $T=100$ grid points.

\end{document}